\documentclass{article}
\usepackage[utf8]{inputenc}
\usepackage{authblk}
\usepackage{setspace}
\usepackage[margin=1.25in]{geometry}
\usepackage{graphicx}
\graphicspath{ {./figures/} }
\usepackage{subcaption}
\usepackage{amsmath}
\usepackage{amsfonts}       
\usepackage{lineno}
\usepackage{times}

\usepackage{custom_commands}
\usepackage{multirow}
\usepackage{booktabs}
\usepackage{array}
\usepackage[table,xcdraw]{xcolor}

\usepackage[style=nejm, 
citestyle=numeric-comp,
sorting=none]{biblatex}
\title{Towards Scalable Foundation Model for\\ Multi-modal and Hyperspectral Geospatial Data}

\author[1]{Haozhe Si\thanks{Correspondence to: Haozhe Si (haozhes3o@illinois.edu) and Han Zhao (hanzhao@illinois.edu)}}
\author[1]{Yuxuan Wan}
\author[1]{Minh Do}
\author[1]{Deepak Vasisht}
\author[1]{Han Zhao}
\author[2]{Hendrik F. Hamann}

\affil[1]{University of Illinois Urbana-Champaign, IL, USA.}
\affil[2]{IBM Research, NY, USA.}

\date{}

\begin{document}

\maketitle

\begin{abstract}
    Geospatial raster (imagery) data, such as that collected by satellite-based imaging systems at different times and spectral bands, hold immense potential for enabling a wide range of high-impact applications. This potential stems from the rich information that is spatially and temporally contextualized across multiple channels (e.g., spectral bands, polarizations) and sensing modalities. Recent work has adapted existing self-supervised learning approaches for such geospatial data. However, they fall short of scalable model architectures, leading to inflexibility and computational inefficiencies when faced with an increasing number of channels and modalities. To address these limitations, we introduce Low-rank Efficient Spatial-Spectral Vision Transformer (LESS ViT) with three key innovations: i) the LESS Attention Block that approximates high-dimensional spatial-spectral attention through Kronecker's product of the low-dimensional spatial and spectral attention components; ii) the Continuous Positional-Channel Embedding Layer that preserves both the continuity and physical characteristics of each spatial-spectral patch; and iii) the Perception Field Mask that exploits local spatial dependencies by constraining attention to neighboring patches. To evaluate the proposed innovations, we construct GFM-Bench, which serves as a comprehensive benchmark for such geospatial raster data. We pretrain LESS ViT using a Hyperspectral Masked Autoencoder framework with integrated positional and channel masking strategies. Experimental results demonstrate that our proposed method achieves competitive performance against state-of-the-art multi-modal geospatial foundation models while outperforming them on cross-satellite generalization tasks with higher computational efficiency. The flexibility and extensibility of our framework make it a promising direction for future geospatial data analysis tasks that involve a wide range of modalities and channels. Code and project page available at \texttt{https://uiuctml.github.io/GeospatialFM/}.
\end{abstract}    


\section{Introduction}
\label{sec:intro}

\begin{figure}
    \centering
    \includegraphics[width=\linewidth]{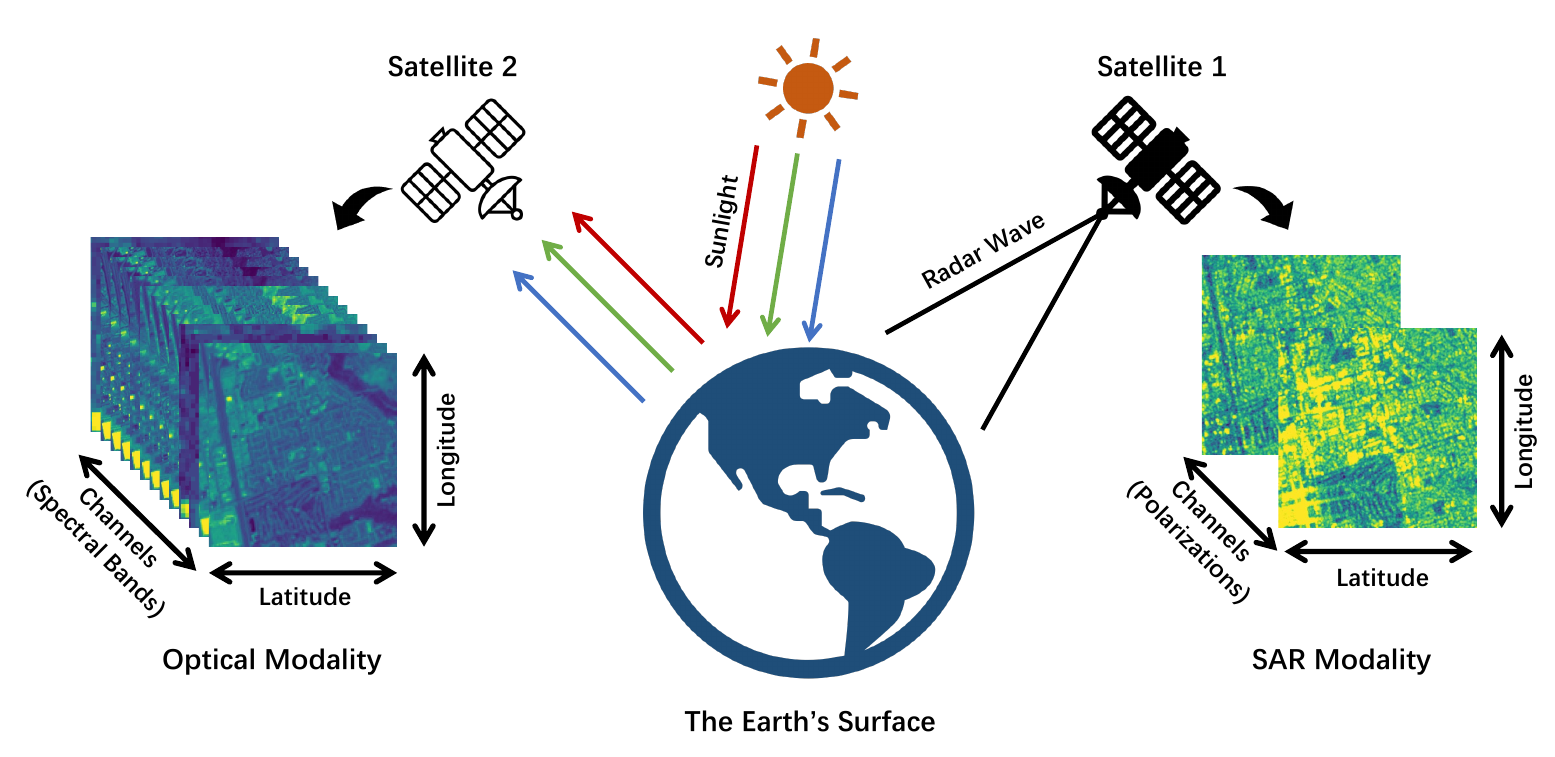}
    \caption{\textbf{Illustration of Multi-modal Geospatial Data.} Two satellite systems capture complementary modalities: optical imagery (left) with multiple spectral bands, and Synthetic Aperture Radar (SAR) imagery (right) with different polarization. Both modalities capture information across spatial and spectral dimensions. While our analysis focuses on optical and SAR data in this work, Earth observation systems can incorporate additional modalities such as thermal infrared and atmospheric measurements. }
    \label{fig:mm_img}
\end{figure}

Geospatial data provides location-specific, timestamped information about the Earth's surface. The rapid development and proliferation of satellite-based imaging systems have led to a significant increase in geospatial raster (e.g., imagery) data collection, offering valuable insights into various aspects of our planet. Geospatial raster data\footnote{Although we focus on static muti-modal and hyperspectral geospatial raster data in this work, we refer in the following as geospatial data for simplicity.} is inherently multi-modal, integrating observations from diverse sensing systems such as optical and radar satellites, as illustrated in \Cref{fig:mm_img}. Each modality captures distinct data dimensions through multiple channels (e.g., spectral bands, polarizations\footnote{We consider SAR polarization channels analogous to spectral bands in this work.}), while introducing complexities from multi-temporal observations, non-ideal imaging conditions, and varying spatial resolutions.

These properties pose significant challenges for interpreting and analyzing geospatial data, requiring most often expert knowledge and specialized tools, which has hampered the utilization of such data. To address these challenges, the remote sensing community has adapted deep-learning methods to process and extract information from geospatial data for specific tasks, including monitoring land use~\cite{crop_landuse, crop_landuse_2, dynamic_world} and climate change~\cite{ghe, ghe2, climate_change}, forecasting natural disasters~\cite{flood, flood2, fire, fire2}, and managing natural resource~\cite{forest, forest2, forest3}. While these task-specific models have demonstrated effectiveness, their reliance on supervised learning approaches requires extensive high-quality labeled data annotated by experts, which is often not available or can be too time-consuming and costly to obtain. Furthermore, the specialized nature of these models limits their applicability to a wider range of problems.

Motivated by the success of foundation models in natural language processing and computer vision~\cite{gpt, dinov2, clip, llava}, one natural direction is to develop large-scale, self-supervised models for geospatial applications. Self-supervised learning (SSL) allows models to benefit from the vast amounts of unlabeled geospatial data and learning useful representations. These pretrained models can then be quickly adapted to specific downstream tasks using smaller labeled datasets, reducing the need for expensive manual annotations. Recent works on geospatial foundation models have attempted to adapt existing SSL paradigms~\cite{mae, coca, dino, dinov2, moco} to geospatial datasets using various strategies. For example, CROMA~\cite{croma} combines reconstruction and contrastive losses to align features from different modalities, SatMAE~\cite{satmae} designs specific patch embedding layers to support multi-temporal and hyperspectral data, and Channel-ViT~\cite{channel_vit} applies attention mechanisms to individual channels of hyperspectral images. Although these approaches achieved empirical success, their underlying architectures and objectives remain largely the same as those designed for natural images\footnote{By natural images, we refer to everyday photographs, e.g., those in conventional computer vision datasets.} and thus do not fundamentally suitable to fully capture the spatial, spatial-channel and inter-channel relations of geospatial data. These relationships are critical for downstream geospatial tasks. For instance, spatial features alone may fail to differentiate aged asphalt roads from water bodies, despite their markedly different spectral characteristics. Therefore, developing model architectures that explicitly encode these distinctive relationships in geospatial data and while efficiently scaling to thousands of spectral channels would significantly advance existing approaches that predominantly leverage spatial features.

To design a framework specific to geospatial data, we highlight three key characteristics that distinguish it from natural images. First, geospatial data, especially land cover observations, exhibits strong spatial autocorrelation that decays with physical distance, in contrast to natural images where correlations are primarily driven by semantic relationships. In this work, we prioritize this distance-dependent correlation by introducing a Perception Field Mask that constrains the spatial attention map, limiting each patch's attention to its local neighborhood when generating spatial representations. 

Second, each pixel in geospatial data corresponds to a specific physical area on Earth with a defined spatial resolution. This intrinsic spatial property makes geospatial data particularly sensitive to resizing operations, as they alter the spatial resolution and complicate model generalization across varying input sizes. To overcome this resolution dependency, we implement a continuous positional embedding that computes inter-patch distances based on physical spatial resolution rather than grid indices. Moreover, the proposed Perception Field Mask facilitates generalization across input resolutions by maintaining a consistent number of attended patches in the attention computation. 

Finally, geospatial data encompasses often many channels  (e.g., multiple spectral bands) but also sensing modalities, such as Multi-Spectral Imagery (MSI) and Synthetic Aperture Radar (SAR). These modalities can be spatially (and temporally) aligned, where for example each channel represents electromagnetic wave reflectance at specific wavelengths and polarizations. To preserve the physical meaning of channels, we implement a continuous channel embedding that encodes channel wavelengths into tokens. To handle varying input channels, we implement a Tied Patch Embedding layer. We further introduce the Low-rank Efficient Spatial-Spectral (LESS) attention block, which replaces standard ViT attention layers to efficiently compute both spatial and spectral attention for geospatial data. The proposed LESS ViT architecture significantly reduces the computational complexity of spatial-spectral attention from $O(N^2C^2)$ to $O(NC)$, where $N$ and $C$ represent the image size and number of spectral channels, respectively. This improved efficiency enables practical scaling to higher spectral dimensionality.

For pretraining, we extend the Masked Autoencoder (MAE) framework to Hyperspectral MAE (Hyper-MAE), specifically designed for hyperspectral data. To standardize evaluation protocols, we construct GFM-Bench with proper validation splits and consistent metrics across diverse geospatial tasks. Extensive experiments demonstrate the effectiveness of our proposed architecture and pretraining strategy. Our contributions are three-fold:

\begin{itemize}[noitemsep,topsep=0pt]
    \item We propose LESS ViT, a novel architecture featuring Tied Patch Embedding, Continuous Positional-Channel Embedding, Perception Field Mask, and LESS attention blocks. This design addresses the characteristics of geospatial data while efficiently exploiting spatial-spectral correlations.
    \item We develop Hyper-MAE, a MAE variant specialized for hyperspectral geospatial data pretraining. By combining masked patch and channel reconstruction, Hyper-MAE introduces a more challenging pretraining objective that encourages learning of inter-channel relationships.
    \item We introduce GFM-Bench, a standardized benchmark for evaluating geospatial foundation models. Our approach demonstrates superior performance against state-of-the-art methods across the benchmark tasks.
\end{itemize}
\section{Related Works}
\textbf{Geospatial Datasets.}~Geospatial datasets provide data with different modalities, covering various spectra and time ranges, depending on the data product. Sentinel-1\&2 are among the most widely used datasets, offering up to 8 radar bands and 13 optical bands, respectively. Landsat~\cite{landsat} provides another valuable source of optical imagery, with up to 11 bands. Previous works have pretrained their models on large classification datasets, such as BigEarthNet~\cite{ben} and fMoW~\cite{fmow}. SeCo~\cite{seco} and SSL4EO~\cite{ssl4eo} propose pretraining datasets specifically designed for SSL methods, which capture both temporal contrast and spatial diversity in geospatial data. Although diverse spectral channels provide rich information for pretraining, the selection of appropriate data types and channels is crucial for different downstream tasks. For instance, multi-temporal data is essential for accurate land cover classification tasks~\cite{africa, pastis, glad}, whereas the first channel of Sentinel-2, which focuses on atmospheric particles, will be dropped in some downstream applications~\cite{so2sat}. Given the diversity in geospatial data, it is necessary to design a flexible model that can handle multi-modal and hyperspectral data effectively.

\textbf{Remote Sensing Representations.}~Researchers have developed large-scale geospatial foundation models (GFMs).These GFMs often employ SSL methods to learn from vast amounts of unlabeled geospatial data. Several SSL techniques have been explored in GFMs, including contrastive learning \cite{simclr, moco, clip, dinov2} and reconstruction-based methods \cite{mae, vae}. In addition to simply applying general SSL methods to geospatial data, recent works also address specific characteristics of the geospatial data, such as hyperspectral \cite{satmae, spectralgpt, ss-mae, channel_vit}, spatiotemporal \cite{satmae, seco,s4}, multi-modal \cite{msgfm, croma, skysense}, varying resolution \cite{scale-mae}, and large volume \cite{gfm} aspects. While these studies aim at exploiting multi-modal and hyperspectral information, their approaches are either computationally inefficient \cite{channel_vit} or parameter-inefficient \cite{ss-mae, croma, skysense}. Furthermore, previous works mainly focus on improving training paradigms to adapt existing ViT architectures to geospatial data. However, ViT is designed for natural images, which typically have only 3 channels and a single modality. Given the gap between the natural image and geospatial data, it is crucial to redesign an architecture that can fundamentally address the inherent characteristics of geospatial data efficiently.
\begin{figure}
    \centering
    \includegraphics[width=0.4\linewidth]{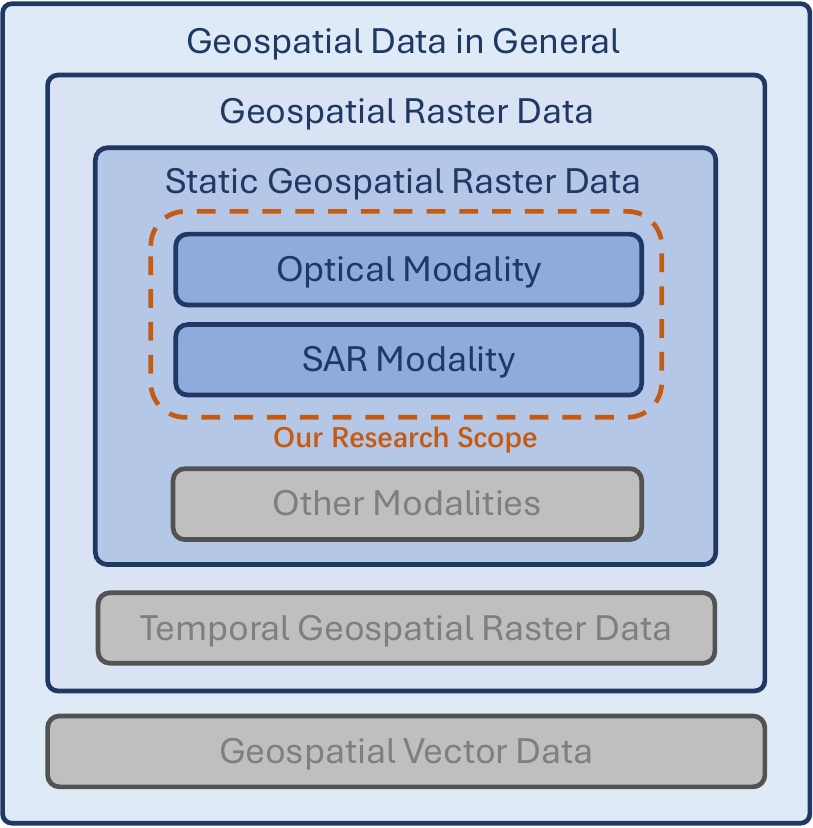}
    \caption{\textbf{Illustration of Research Scope.} In this work, we focus on static geospatial raster data in optical and SAR modalities and refer them as geospatial data. However, geospatial data in general can also exist as vector data, incorporate temporal dimensions, and span various other modalities.}
    \label{fig:term}
\end{figure}

\section{Terms of Terminology}
In this work, we propose an architecture and framework for multi-modal and hyperspectral geospatial datasets. 
Given the diverse terminology in this paper, we first clarify key concepts.

\textbf{Geospatial Data.} Geospatial data encompasses both raster and vector formats. These formats can be further categorized as temporal (containing a time dimension) or static (without temporal information). Static geospatial raster data can be classified by modality, with our work primarily focusing on optical and SAR modalities. Figure~\ref{fig:term} illustrates the hierarchical organization of our research scope. Our current study focuses on static geospatial datasets, while extension to temporal geospatial data is planned for future work.

\textbf{Modality.} In our context, modalities are differentiated by their image acquisition methods. For example, optical images are acquired passively by collecting surface reflections across various spectral bands, while SAR (Synthetic Aperture Radar) images are actively acquired by emitting radar waves toward the ground and measuring the reflected signals with different polarizations. 

\textbf{Dimension.} We refer to spatial, spectral, and temporal characteristics as different dimensions. For instance, a static optical geospatial data $x\in\mathbb{R}^{C\times N}$ has two dimensions, where $C$ represents the spectral dimension comprising concatenated spectral bands from optical imagery and polarizations from SAR imagery, and $N$ represents the flattened spatial dimension combining longitude and latitude. In contrast, a temporal optical geospatial data $x\in\mathbb{R}^{T\times C\times N}$ includes an additional temporal dimension $T$, corresponding to the number of sequential frames in the data sample. Note that we also use embedding dimension to denotes the number of features utilized in our model architectures. It is important to differentiate this usage of dimensions from the data dimensions (spatial, spectral, temporal, etc.) described above.

\textbf{Channels and Hyperspectral.} In this paper, we sometimes use the term ``channel'' interchangeably with spectral dimension. We use ``hyperspectral'' to indicate that our data contains significantly more channels than standard RGB images, which typically have only three channels.

\section{Low-rank Efficient Spatial-Spectral ViT}
In this section, we introduce our Low-rank Efficient Spatial-Spectral (LESS) ViT architecture. Specifically, we elaborate the three key components of the framework: the hyperspectral patch embedding block, the LESS Attention Block and the Perception Field Mask.

\subsection{Hyperspectral Patch Embedding} \label{sec:embedding}
Hyperspectral images can contain tens to thousands of channels, distinguishing them from natural images that typically have three (RGB) channels. The rich spectral information encoded in these channels can exhibit strong physical correlations that must be effectively leveraged. To exploit these spectral dependencies in subsequent attention blocks, we adopt a Tied Patch Embedding Layer that maintains spectral fidelity by explicitly embedding each channel's information, and incorporate a continuous positional-channel embedding to capture both spatial and spectral relationships.

\begin{figure}[tb]
    \centering
    \includegraphics[width=0.8\columnwidth]{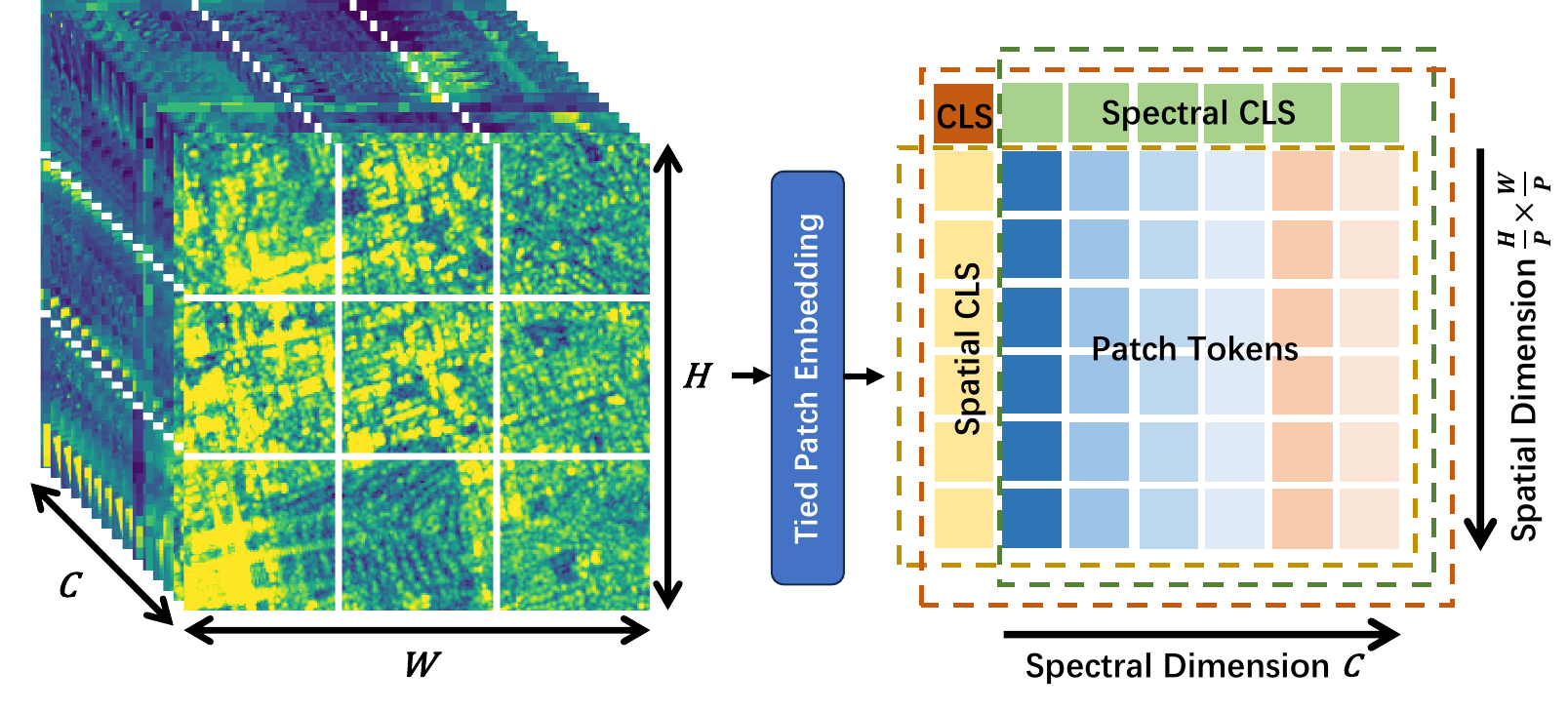}
    \caption{\textbf{Hyperspectral Patch Embedding.} Hyperspectral images, with dimensions $C\times H \times W$, are embedded into spatial-spectral tokens through the Tied Patch Embedding Layer. We then prepend the Spatial, Spectral and global \texttt{[CLS]} tokens to the resulting patch tokens. Among which, the Spatial \texttt{[CLS]} tokens (yellow box) represent every spatial patch across the spectrum, Spectral \texttt{[CLS]} (green box) tokens represent the information of every channel and the global \texttt{[CLS]} token (red box) is the representation of all the spatial-spectral tokens.}
    \label{fig:hpe}
\end{figure}

\textbf{Tied Patch Embedding Layer.}~Given a hyperspectral image with dimensions $C\times H \times W$, where $C$ denotes the number of channels, the tied patch embedding layer~\cite{channel_vit} partitions the image into $C\times \frac{H}{P}\times \frac{W}{P}$ patches of size $P\times P$. A shared learnable projection matrix $W\in \mathbb R^{P^2 \times D}$ transforms patches from each channel into $D$-dimensional tokens. This weight-sharing mechanism across channels~\cite{shared_pe} ensures channel-independence, making the architecture adaptable to geospatial data with varying spectral dimensions. The resulting spatial-spectral tokens have dimension $\mathbb R^{N\times C\times D}$, where $N= \frac{H}{P}\times \frac{W}{P}$.

To capture global context, we augment these tokens with specialized \texttt{[CLS]} tokens: (1) $N$ Spatial \texttt{[CLS]} tokens are prepended along each spectral slice to summarize patch information at each spatial location, (2) $C$ Spectral \texttt{[CLS]} tokens are prepended along the spatial dimension to aggregate spectral information, and (3) a global \texttt{[CLS]} token is prepended to all tokens. This results in a final token structure of shape $\mathbb R^{(N+1)\times (C+1)\times D}$, as illustrated in \Cref{fig:hpe}. To account for the distinct data distribution of optical and radar imagery, we employ modality-specific Tied Patch Embedding Layers to extract spatial-spectral tokens. The resulting token representations are concatenated and processed by subsequent network modules.

\begin{figure}
    \centering
    \includegraphics[width=0.9\linewidth]{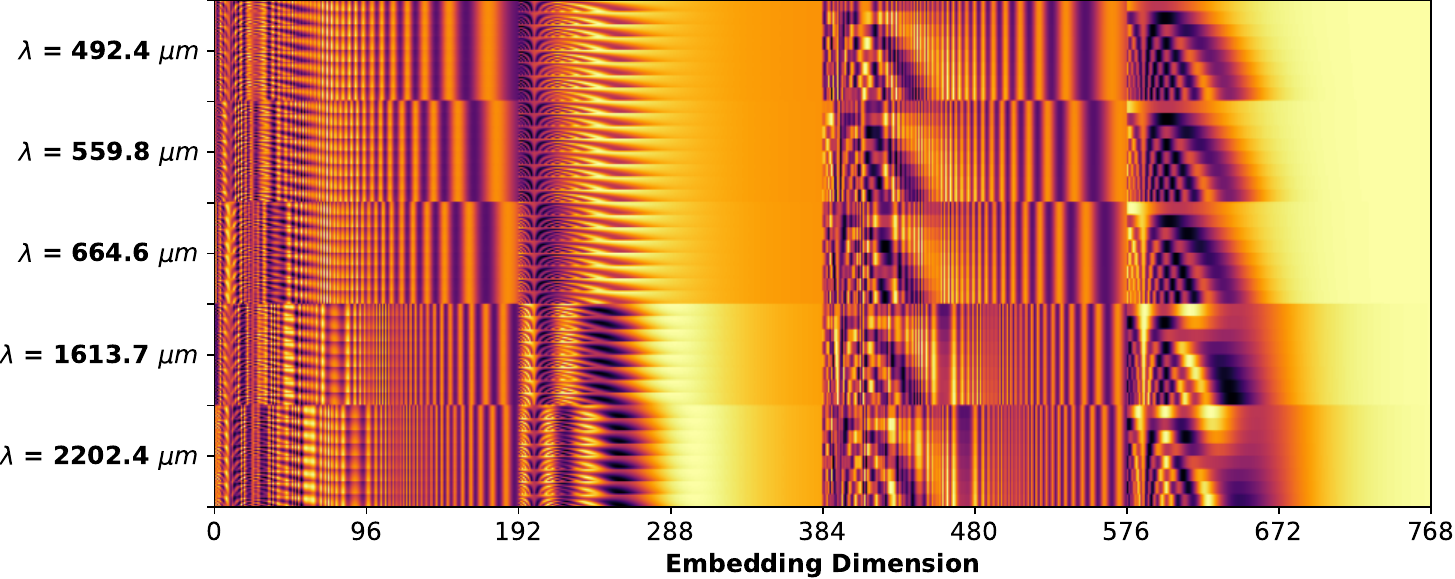}
    \caption{\textbf{Continuous Positional-Channel Embedding.} The visualization demonstrates the embedding patterns across spatial positions and spectral bands ($\lambda$). For each wavelength, the embedding exhibits continuous spatial variation, while the overall pattern evolves systematically with increasing wavelength. The embedding encodes both spatial positions and spectral information in a unified representation.
    }
    \label{fig:pce}
\end{figure}

\textbf{Continuous Positional-Channel Embeddings.}~Unlike natural images where pixel positions are purely relative, pixels in geospatial data represent physical locations on Earth's surface, characterized by their spatial resolution (e.g., meters per pixel). While different datasets may have varying spatial resolutions, we aim to create a unified positional representation that captures absolute distances between patches regardless of the underlying pixel resolution. To achieve this, we modify the conventional sinusoidal positional embedding by incorporating both spatial resolution and patch size to compute actual geographic distances between patches:
\begin{equation}
\begin{aligned}
    &\mathrm{PE}_{(x,r,p, 2i)} = \sin(xrp/10000^{2i/d}),\\
    &\mathrm{PE}_{(x,r,p, 2i+1)} = \cos(xrp/10000^{2i/d}),
\end{aligned}
\end{equation}
where $x$ denotes the grid index, $p$ is the patch size, $r$ represents the image spatial resolution in meters per pixel, $d$ is the model's embedding dimension, and $i \in \{0, 1, \dots, \lfloor d/2 \rfloor - 1\}$. 

Each hyperspectral channel corresponds to a specific band, capturing electromagnetic radiation at a distinct wavelength. To leverage the channel structure preserved by our Tied Patch Embedding Layer, we encode spectral information using continuous channel embeddings based on physical wavelengths:
\begin{equation}
\begin{aligned}
&\mathrm{PE}{(\lambda, 2i)} = \sin(\lambda/10000^{2i/d}),\\
&\mathrm{PE}{(\lambda, 2i+1)} = \cos(\lambda/10000^{2i/d}),
\end{aligned}
\end{equation}
where $\lambda$ denotes the central wavelength of each band. This physics-informed embedding enables the model to capture meaningful inter-channel relationships and generalize to arbitrary spectral bands, as it maps channels to a continuous spectral space rather than treating them as discrete indices. Finally, we sum the spatial and spectral embeddings to form our continuous positional-channel embedding that jointly encodes both geographic distances and spectral wavelength, as illustrated in \Cref{fig:pce}.

\subsection{Low-rank Efficient Spatial-Spectral Attention}\label{sec:less}
\begin{figure}
    \centering    
    \includegraphics[width=0.3\columnwidth]{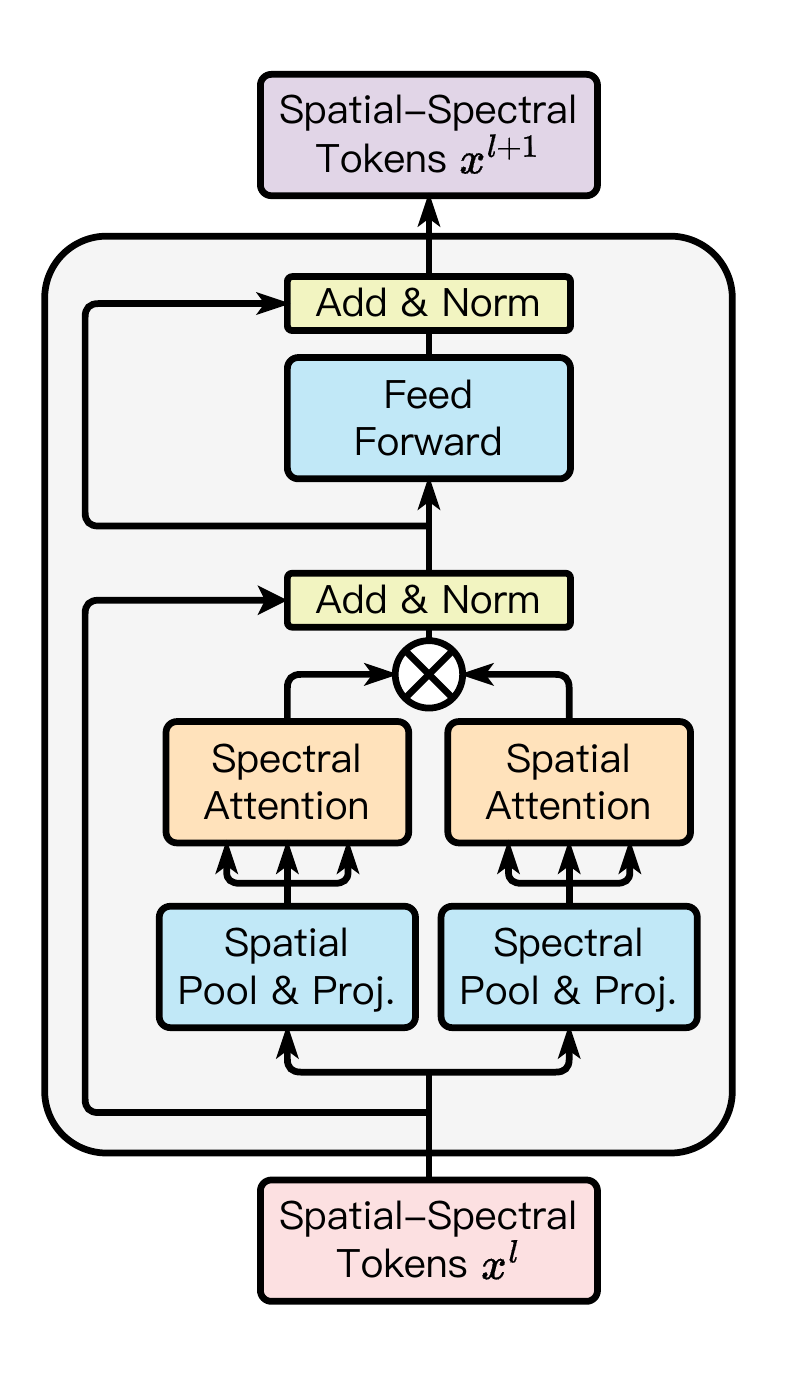}
    \caption{\textbf{LESS Attention Block.} An illustration of the LESS Attention Block, which decomposes spatial and spectral attention computations and approximates the full spatial-spectral attention through a Kronecker product of the individual attention maps. Multiple LESS Attention Blocks connected in series with each block transforming input $x^{l}$ to produce the corresponding output $x^{l+1}$ for the subsequent layer.
    }
    \label{fig:less-vit}
\end{figure}

Given the spatial-spectral tokens $X\in\mathbb R^{N\times C\times D}$, we would like to apply the attention mechanism to capture the correlations of the spatial and spectral patches. A straightforward yet inefficient approach~\cite{channel_vit} is to flatten the spatial-spectral tokens into $\bar X\in\mathbb R^{NC \times D}$ and apply the standard attention mechanism. The computational complexity of this approach scales quadratically with the number of channels and tokens $(O(N^2C^2))$, making it infeasible for geospatial data with a large number of channels or tokens. To address this limitation, we propose the Low-rank Efficient Spatial-Spectral Vision Transformer (LESS ViT). LESS ViT consists of multiple LESS attention blocks specifically designed for spatial-spectral tokens. An illustration of the blocks is shown in \Cref{fig:less-vit}. The computation complexity of LESS ViT is reduced to $O(NC)$, which scales linearly with the number of spatial-spectral tokens. 

\setlength{\textfloatsep}{5pt}
\begin{algorithm*}[tb]
    \caption{Low-Rank Efficient Spatial-Spectral (LESS) Attention Block}\label{algo:low-rank-spatial-spectral}
    \begin{algorithmic}[1]
        \Require Input tokens $X \in \mathbb{R}^{N \times C \times D}$, Sub-dimensions $d_1d_2=D$
        \Ensure Output tokens $Y \in \mathbb{R}^{N \times C \times D}$
        \State \textbf{\# Decompose spatial-spectral tokens}
        \State $X_S \gets \textsc{AttenPool}(X, \text{dim}=1)\in \mathbb{R}^{N \times d_1}$ \Comment{Spatial-only tokens}
        \State $X_C \gets \textsc{AttenPool}(X, \text{dim}=0)\in \mathbb{R}^{C \times d_2}$ \Comment{Spectral-only tokens}
        \State \textbf{\# Compute spatial-only attention}
        \State $Q_S, K_S, V_S \gets X_S W_{Q_S}, X_S W_{K_S}, X_S W_{V_S}\in \mathbb{R}^{N \times d_1}$
        \State $Y_S \gets \textsc{Softmax}(Q_S K_S^\top/\sqrt{d_1})V_S\in \mathbb{R}^{N \times d_1}$ \Comment{Complexity $O(N^2d_1)$}
        \State \textbf{\# Compute spectral-only attention}
        \State $Q_C, K_C, V_C \gets X_C W_{Q_C}, X_C W_{K_C}, X_C W_{V_C}\in \mathbb{R}^{C \times d_2}$
        \State $Y_C \gets \textsc{Softmax}(Q_C K_C^\top/\sqrt{d_2})V_C\in \mathbb{R}^{C \times d_2}$ \Comment{Complexity $O(C^2d_2)$}
        \State \textbf{\# Approximate the spatial-spectral attention}
        \State $Y \gets Y_C\otimes Y_S\in \mathbb{R}^{NC \times d_1d_2}$ \Comment{Complexity $O(NCD)$}
        \State $Y \gets \textsc{Reshape}(Y, (N, C, D))$
        \State \Return $Y$
    \end{algorithmic}
\end{algorithm*}

To efficiently model spatial-spectral interactions, our LESS attention block approximates the full spatial-spectral attention matrix using a Kronecker product of separate spatial and spectral attention matrices. As shown in \Cref{algo:low-rank-spatial-spectral}, the block first decomposes input tokens $X$ into spatial tokens $X_S$ and spectral tokens $X_C$ through an Attention Pooling Layer (\textsc{AttenPool}). This layer pools the tokens sequence along one dimension by computing its low-dimensional representations: it uses corresponding \texttt{[CLS]} tokens to generate queries, while deriving keys and values from all tokens. The resulting attention output maintains the same dimensionality as the \texttt{[CLS]} tokens before being projected to the target dimension through a linear layer.

Then, the spatial attention matrix $A_S$ and the spectral attention matrix $A_C$ are calculated separately using $X_S$ and $X_C$, along with their respective value matrices $V_S$ and $V_C$. Since $A_S$ and $A_C$ represent convex combinations of spatial and spectral dimensions respectively, their Kronecker product $A = A_C \otimes A_S$ yields a convex combination over the joint spatial-spectral dimensions. Leveraging the mixed-product property, we efficiently obtain a low-rank approximation of the full attention computation:

\begin{equation}
    \begin{aligned}
        Y = AV \coloneqq \sum_{i=1}^r(A_C^i \otimes A_S^i)(V_C \otimes V_S)
               = \sum_{i=1}^r(A_C^iV_C) \otimes (A_S^iV_S)
               = \sum_{i=1}^rY_C^i \otimes Y_S^i,
    \end{aligned}
    \label{eq:less}
\end{equation}
where $Y_S^i = A_S^iV_S \in \mathbb{R}^{N \times d_1}$ and $Y_C^i = A_C^iV_C \in \mathbb{R}^{C \times d_2}$, $\forall i\in\{1,\cdots r\}$. This approach avoids explicitly constructing the full attention matrix $A$, thereby reducing computational complexity. Note that $r$ in \Cref{eq:less} is a hyperparameter for rank control. By adjusting the rank, we can increase the capacity of the attention map while maintaining the same computational complexity order. 

The LESS attention block offers significant efficiency advantages compared to the previous spatial-spectral attention approach~\cite{channel_vit}, which directly applies attention to the reshaped tokens $\bar{X} \in \mathbb{R}^{NC \times D}$. It reduces the computational complexity from $O(N^2C^2D)$ to $O(rN^2d_1 + rC^2d_2+rNCD)$, where $d_1d_2 = D$ and $r \ll \min(N, D)$ is a small constant. In practice, more dimensions are assigned to $d_1$ for spatial features and fewer to $d_2$ for spectral features, as spectral correlation is easier to learn. This approximation makes the attention block more scalable to larger numbers of channels while still capturing spatial-spectral interactions.

\subsection{Perception Field Mask}
\begin{figure}[t]
    \centering
    \includegraphics[width=0.8\columnwidth]{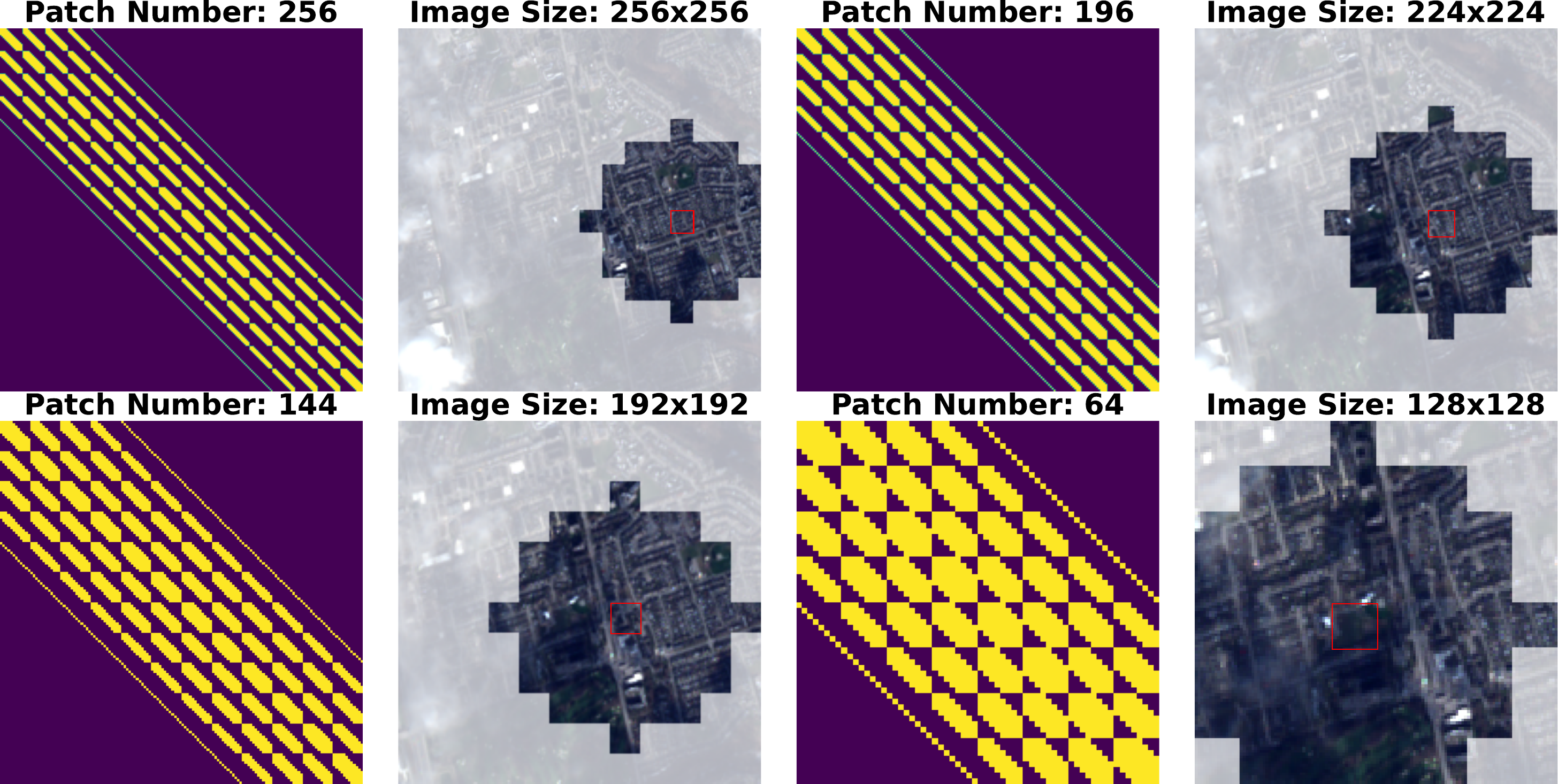}
    \caption{\textbf{Perception Field Mask.} Visualization of Perception Field Masks for varying image size and their impact on attention computation. Despite varying image sizes and patch numbers, the attention region for the patch highlighted by the red box consistently covers the same spatial area in the input image. This distance-based masking mechanism enables uniform attention patterns across different resolutions without requiring input resizing.}
    \label{fig:pmf}
\end{figure}

Tobler's first law of geography~\cite{tobler} states that:
\begin{quote}
\textit{``Everything is related to everything else, but near things are more related than distant things.''}
\end{quote}
To explicitly model spatial autocorrelation in geospatial data, we introduce the Perception Field Mask, as illustrated in \Cref{fig:pmf}. The Perception Field Mask constrains the spatial attention computation by allowing each token to attend only to patches within a specified distance threshold. We define this threshold in meters rather than pixels, ensuring consistent spatial relationships across different image resolutions. While our approach emphasizes local spatial relationships, we acknowledge that geospatial data can exhibit meaningful long-range dependencies. For instance, in global climate studies, atmospheric patterns or ocean currents demonstrate significant interactions across continental scales. Recognizing this, our Perception Field Mask introduces a hyperparameter that controls the scale of locality relevant to the downstream task, allowing flexible modeling of spatial relationships. Consequently, This distance-based masking mechanism offers two key advantages: (1) it provides a tunable hyperparameter to control the locality of attention computation, aligning with Tobler's law, and (2) it enables the model to process images of varying sizes without downsampling, as the attention field remains spatially consistent regardless of resolution. 

Through the combination of our Hyperspectral Patch Embedding module and the Perception Field Mask, LESS ViT achieves flexible processing of geospatial data with arbitrary spectral channels and spatial dimensions. This architecture eliminates the need for input resizing or model retraining, enabling direct analysis of hyperspectral data at their native resolutions. As shown in \Cref{tab:summarize}, these architectural innovations overcome key limitations of existing approaches.

\begin{table}[t]
    \centering
    \caption{\textbf{Comparision with ViT~\cite{vit} and Channel-ViT~\cite{channel_vit}.} Our proposed LESS ViT provides a more flexible and physics-informed architecture while maintaining computational efficiency.}
    \label{tab:summarize}
    \footnotesize
    \renewcommand{\arraystretch}{1.2} 
    \setlength{\tabcolsep}{1pt} 
    \resizebox{\textwidth}{!}{
\begin{tabular}{l@{\hspace{15pt}}c@{\hspace{15pt}}c@{\hspace{15pt}}c@{\hspace{15pt}}c}
\toprule
\textbf{Architecture} & \textbf{Input Channel Flexibility}   & \textbf{Preserves Native Resolution} & \textbf{Embedding Mechanism} & \textbf{Computation Efficiency} \\ \midrule
ViT~\cite{vit}          & \xmark                      & \xmark                      & Pixel-based         & High                   \\
Channel-ViT~\cite{channel_vit}  & \cmark                      & \xmark                      & Pixel-based         & Low                    \\
LESS ViT     & \cmark                      & \cmark                      & Physics-informed    & Medium   \\ \bottomrule
\end{tabular}
}
\end{table}

\section{Hyperspectral Masked Autoencoder}\label{sec:hyper-mae}
We extend the Masked Autoencoder (MAE) framework to effectively handle hyperspectral geospatial data. While recent works~\cite{satmae, spectralgpt, s2mae} have adapted MAE for pretraining geospatial foundation models, they all employ uniform random masking across both spatial and spectral dimensions. This approach, which masks tokens with equal probability before encoding and reconstructs all masked patches during decoding, potentially oversimplifies the pretraining objective. Specifically, models can exploit cross-channel spatial redundancies without fully capturing the intrinsic spatial-spectral correlations in the data.

\begin{figure}[tb]
    \centering
    \includegraphics[width=0.8\linewidth]{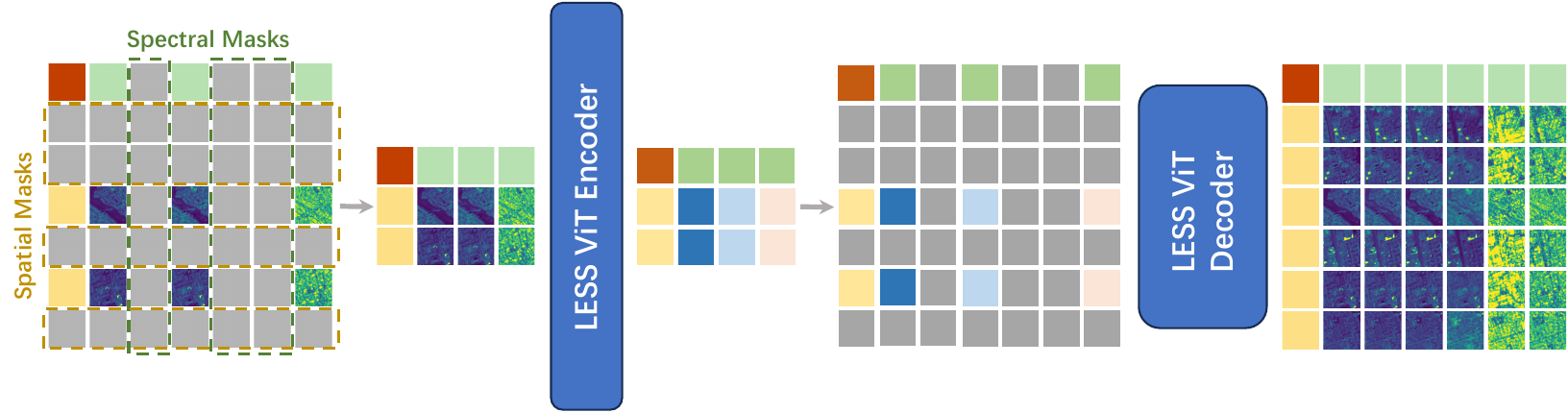}
    \caption{\textbf{Hyper-MAE.} Hyper-MAE employs a LESS ViT encoder-decoder architecture. The framework incorporates decoupled spatial and spectral masking to create a more challenging self-supervised pretraining objective.}
    \label{fig:hypermae}
\end{figure}

To address this limitation, we propose Hyperspectral Masked Autoencoder (Hyper-MAE), which decouples spatial and spectral masking. An illustration is shown in \Cref{fig:hypermae}. During training, we randomly mask 75\% of spatial patches and 50\% of spectral channels before encoding. The decoder then reconstructs the complete hyperspectral image at pixel level from these partially observed tokens. Importantly, we apply identical spatial masks across all unmasked channels, similar to the tube masking strategy in~\cite{videomae}. This approach encourages the model to learn intrinsic spatial and spectral correlations in geospatial data, as it cannot rely on positional information from other channels to reconstruct masked spatial patches.

Hyper-MAE adopts the LESS ViT-based architectures as the encoder and the lightweight decoder. Following MAE, we compute the mean squared error (MSE) between the reconstructed and original images at the pixel level, focusing only on masked regions. To independently optimize spatial and spectral reconstruction, we decompose the loss into two components: 
\begin{equation}
    \mathcal{L}_{\mathrm{spatial}} = \frac{1}{|M|} \sum_{i\in M} \left( \hat{I}^i - I^{i} \right)^2,
\end{equation}
where $M$ is the set of pixels belonging to the masked spatial positions. $\hat I$ is the reconstructed images, and
\begin{equation}
    \mathcal{L}_{\mathrm{spectral}} = \frac{1}{|N|} \sum_{j\in N} \left( \hat{I}_j - I_{j} \right)^2,
\end{equation}
where $N$ is the set of pixels in the masked channels. Notably, pixels that are masked in both spatial and spectral dimensions contribute to both loss terms. This intentional double-counting encourages the model to pay special attention to these challenging regions where both spatial and spectral information is missing.

Similar to MAE, spatial token reconstruction encourages the learning of spatial dependencies, leading to robust spatial representations. Complementing this spatial learning, our spectral masking strategy brings multiple key benefits. Most importantly, it drives the model to learn intrinsic spectral correlations that reflect the physical properties of observed objects. Second, it enables adaptation to diverse spectral signatures and facilitates generalization across channel configurations. Finally, from an implementation perspective, processing fewer channels during training naturally leads to improved computational efficiency.

\section{GFM-Bench}
\begin{table}[t]
\centering
\caption{\textbf{GFM-Bench.} Characteristics of datasets in the benchmark. All of the datasets are open-source datasets. For datasets without predefined validation splits~\cite{so2sat, spectralgpt, dfc}, we reserve 10\% of the training data for validation.}
\label{table:geo-bench}
\resizebox{\textwidth}{!}{%
\begin{tabular}{@{}llccccccccc@{}}
\toprule
\textbf{Type} & \textbf{Name} & \textbf{Image Size} & \textbf{\# Classes} & \textbf{Train} & \textbf{Val} & \textbf{Test} & \textbf{\# Bands} & \textbf{Resolution} & \textbf{Sensors}  \\ \midrule
\multirow{3}{*}{Classification} 
& BigEarthNet~\cite{ben}  & 120 $\times$ 120 & 19 & 269,695 & 123,723 & 125,866 & 12/2 & 10.0 & Sentinel-1/2\\
& So2Sat~\cite{so2sat}      & 32 $\times$ 32  & 17 & 31,713 & 3,523 & 48,307 & 10 & 10.0 & Sentinel-2  \\
& EuroSat~\cite{eurosat}      & 64 $\times$ 64  & 10 & 16,200 & 5,400 & 5,400 & 13 & 10.0 & Sentinel-2 \\ \midrule
\multirow{4}{*}{Segmentation} 
& SegMunich~\cite{spectralgpt}   & 128 $\times$ 128 & 13  & 3,000 & 403 & 403 & 10 & 10.0 & Sentinel-2 \\
& DFC2020~\cite{dfc} & 96 $\times$ 96 & 8  & 41,537 & 4,615 & 8,874 & 13 & 10.0 & Sentinel-2  \\
& MARIDA~\cite{marida}    & 96 $\times$ 96 & 11  & 5,622 & 624 & 6,183 & 11 & 10.0 & Sentinel-2  \\
& NLCD-L~\cite{ssl4eo-l}  & 128 $\times$ 128 & 21 & 17,500 & 3,750 & 3,750 & 20 & 30.0 & Landsat 7-9 \\ \bottomrule
\end{tabular}%
} 
\end{table}

Several datasets~\cite{so2sat, dfc, spectralgpt, ben} do not include validation sets, forcing prior work to tune hyperparameters on test data. To address this limitation, GEO-Bench~\cite{geobench} establishes a rigorous evaluation framework for geospatial foundation models. Following this precedent, we introduce GFM-Bench, implemented using the HuggingFace\footnote{\texttt{https://huggingface.co/}} framework for ease of use and providing standardized evaluation protocols. The current version of GFM-Bench consists three classification tasks (EuroSAT~\cite{eurosat}, BigEarthNet~\cite{ben} and So2Sat~\cite{so2sat} and four segmentation tasks (SegMunich~\cite{spectralgpt}, DFC2020~\cite{dfc}, MARIDA~\cite{marida}, NLCD-L~\cite{ssl4eo-l}). All datasets are derived from data of the Sentinel constellation except NLCD-L, which uses Landsat data. For datasets without validation splits, we either allocate 10\% of the training data for validation~\cite{so2sat, dfc, spectralgpt} or utilize alternate versions that include validation sets~\cite{ben}. We implement consistent evaluation metrics to ensure fair hyperparameter selection across models. GFM-Bench enforces hyperparameter tuning on validation sets and performance reporting on test sets. An overview of the benchmark is shown in \Cref{table:geo-bench}.

\textbf{BigEarthNet.}~\cite{ben} BigEarthNet is a multi-label classification dataset consisting of 12 MSI bands and 2 SAR bands. The complete training set, validation set and test set we use consist of 269,695 samples, 123,723 samples and 125,866 samples respectively, and we only use 10\% of the samples for fine-tuning and 10\% of the samples for validation. Notice that we adopt the dataset from TorchGeo~\cite{torchgeo}, which provides a different data split from the one used in previous works~\cite{croma} (354,196 training samples and 118,065 test samples). All images in the BigEarthNet dataset are 120 $\times$ 120 pixels.

\textbf{So2Sat.}~\cite{so2sat} So2Sat is a local climate zone (LCZ) classification task consisting of 10 MSI bands. We reserve 10\% of the training set as the validation set, resulting in 31,713 training samples, 3,523 validation samples and 48,307 test samples. Same as BigEarthNet, we use 10\% of the complete training set and validation set during fine-tuning. Images are 32 $\times$ 32 pixels.

\textbf{EuroSAT.}~\cite{eurosat} EuroSAT is a land cover and land use classification dataset containing 13 MSI bands. It consists of 16,200 training samples, 5,400 test samples and 5,400 validation sample. Images of EuroSAT are 64 $\times$ 64 pixels.

\textbf{SegMunich.}~\cite{spectralgpt} SegMunich is a land use and land cover segmentation dataset consisting of 10 MSI bands. This is a new Semantic Segmentation dataset collected by~\cite{spectralgpt}. Images of SegMunich are 128 $\times$ 128 pixels.

\textbf{DFC2020.}~\cite{dfc} DFC2020 is a dataset that includes 13 MSI bands. The original DFC2020 dataset contains only a validation set of 986 samples and a test set of 5,128 samples. To better utilize this dataset, we treat the original test set (5,128 samples) as our training and validation sets, and the original validation set (986 samples) as our test set. In addition, since the image resolution is 256 $\times$ 256 pixels, we follow CROMA's method, further dividing each image of 256 $\times$ 256 pixels into 9 smaller patches of 96 $\times$ 96 pixels with the overlap of 16 pixels. As a result, our final training set contains 41,537 training samples, the final validation set contains 4,615 samples and the final test set consists of 8,874 samples. All images are 96 $\times$ 96 pixels.

\textbf{MARIDA.}~\cite{marida} MARIDA is a dataset for sparsely labeled marine debris which consists of 11 MSI bands. This dataset contains a training set of 694 samples along with a validation set of 328 samples and a test set of 350 samples. All image samples are originally 256 $\times$ 256 pixels. We combine both the original validation set and test set into one single test set (678 samples). We employ the same approach as DFC2020's where we divide 256 $\times$ 256 pixels into 9 smaller patches of 96 $\times$ 96 pixels. Thus, our final training set contains 5,622 training samples, 624 validation samples and 6,183 test samples. All images are 96 $\times$ 96 pixels.

\textbf{NLCD-L.}~\cite{ssl4eo-l}~NLCD-L combines optical data from Landsat-7 and Landsat 8-9~\cite{landsat} with NLCD ground-truth labels~\cite{nlcd}, originally proposed in SSL4EO-L~\cite{ssl4eo-l}. The dataset contains 20 MSI bands, deliberately exceeding Sentinel-2's channel count. It comprises 17,500 training samples, 3,750 validation samples, and 3,750 test samples. While the original images are 264 $\times$ 264 pixels, we center-crop to 128 $\times$ 128 to reduce GPU memory requirements during evaluation. In addition to the constructed NLCD-L, GFM-Bench also incorporates the original Landsat benchmarks from SSL4EO-L~\cite{ssl4eo-l}.

\section{Experiments}
We pretrain a LESS ViT-Base model using Hyper-MAE on the SSL4EO-S12 dataset~\cite{ssl4eo}. To evaluate our model, we establish GFM-Bench, a benchmark suite derived from standard geospatial datasets. We present quantitative experimental results to demonstrate out competitive performance against state-of-the-art approaches. 

\subsection{Pretrain and Evaluation Setup}
\textbf{LESS ViT Configuration.}~We take the similar configuration as ViT-Base. Specifically, Hyper-MAE consists of 12 LESS attention blocks for encoder and 8 for decoder, with embedding dimension being 768 and 512 respectively. Each attention block uses 12 heads. The LESS attention blocks employ rank $r=1$ and attention ratio $\frac{d_1}{d_2} = 16$, where $d_1$ is number of dimensions assigned to spatial attention and $d_2$ is the one for spectral attention. The patch size is 16.

\textbf{Pretrain.}~Our LESS ViT is pretrained on the SSL4EO-S12~\cite{ssl4eo} dataset, which is a large-scale multi-modal geospatial dataset containing spatially and temporally aligned MSI with 13 channels and SAR imagery with 2 channels. We concatenate MSI and SAR channels to create a unified hyperspectral input, applying spatial masking to 75\% of patches and channel masking to 50\% of spectral bands. The model is pretrained for 300 epochs using 8 NVIDIA A100 40GB GPUs. Additional pretraining details are provided in \Cref{appendix:pretrain}.

\textbf{Baselines.}~We compare LESS ViT-Base against SoTA geospatial representation learning methods using ViT-Base backbones, including SatMAE~\cite{satmae}, CROMA~\cite{croma}, and SpectralGPT~\cite{spectralgpt}. We also evaluate against ViT-Large models: Scale-MAE~\cite{scale-mae} and SatMAE++~\cite{satmaepp}. All experiments are conducted on GFM-Bench following our standardized evaluation protocol. Additional details are provided in \Cref{appendix:evaluation}.

\subsection{Quantitative Results}\label{sec:quant_res}
\subsubsection{Hyperspectral Optical Experiments}
\begin{table*}[t]
    \centering
    \caption{\textbf{Quantitative results on seven benchmarks under Fine-tuning (FT) and Linear Probing (LP).} We report Top 1 accuracy for classification tasks, mean Average Precision (mAP) for multi-label classification tasks, and mean Intersection over Union (mIoU) for segmentation tasks. * indicates only 10\% of the training and validation sets are used, following previous works. We also report the average performance of each method on all benchmarks. \textbf{Bold} and \underline{underlined} values indicating the highest and second-highest results, respectively. The bottom two rows show ViT-Large models, which serve as references and are not directly compared with the ViT-Base approaches.}
    \label{tab:quant}
    \footnotesize
    \renewcommand{\arraystretch}{1.2} 
    \setlength{\tabcolsep}{1pt} 
    \resizebox{\textwidth}{!}{
    \begin{tabular}{c
                    c
                    >{\centering\arraybackslash}p{1.6cm}
                    >{\centering\arraybackslash}p{1.6cm}
                    >{\centering\arraybackslash}p{1.6cm}
                    >{\centering\arraybackslash}p{1.6cm}
                    >{\centering\arraybackslash}p{1.6cm}
                    >{\centering\arraybackslash}p{1.6cm}
                    >{\centering\arraybackslash}p{1.6cm}
                    >{\centering\arraybackslash}p{1.6cm}
                    >{\centering\arraybackslash}p{1.6cm}
                    >{\centering\arraybackslash}p{1.6cm}
                    >{\centering\arraybackslash}p{1.6cm}
                    >{\centering\arraybackslash}p{1.6cm}}
        \toprule
                                &               & \multicolumn{2}{c}{EuroSAT}       & \multicolumn{2}{c}{BigEarthNet*}  & \multicolumn{2}{c}{So2Sat*}       & SegMunich     & DFC2020       & MARIDA        &               \\
                                &               & \multicolumn{2}{c}{Top 1 Acc.}    & \multicolumn{2}{c}{mAP}           & \multicolumn{2}{c}{Top 1 Acc.}    & mIoU          & mIoU          & mIoU          & \textit{Avg.} \\                  \cmidrule(lr){3-4} \cmidrule(lr){5-6}\cmidrule(lr){7-8}\cmidrule(lr){9-9}\cmidrule(lr){10-10}\cmidrule(lr){11-11}
Method                          & Backbone      & FT            & LP                & FT            & LP                & FT            & LP                & FT            & FT            & FT            &               \\
SatMAE~\cite{satmae}            & ViT-B         & \underline{98.78}& \textbf{96.04}             & 85.84         & 78.69             & \underline{64.97}         & 64.65            & \textbf{44.87}         & \textbf{52.84}         & \underline{54.33}         & \textbf{71.22}         \\
CROMA~\cite{croma}              & ViT-B         & \textbf{98.83}& \underline{95.87}             & \textbf{87.57}         & \textbf{84.90}            & \textbf{66.53}         & \textbf{65.04}             & 39.61         & \underline{49.48}         & 43.04         & 70.10         \\
SpectralGPT~\cite{spectralgpt}  & ViT-B         & 97.94         & 90.57             & 83.78         & 73.29             & 61.63         & 57.49             & \underline{44.73}         & 48.23         &    44.72         & 66.93         \\
\bluerow Ours                   & LESS ViT-B    & 98.06         & 95.12             & \underline{86.08}         & \underline{82.94}             & 63.25         & \underline{64.66}             & 42.29         & 45.60         & \textbf{55.64}         & \underline{70.40}         \\ \midrule
\gray{Scale-MAE~\cite{scale-mae}}& \gray{ViT-L} & \gray{98.78}  & \gray{96.41}      & \gray{84.66}  & \gray{73.69}      & \gray{66.48}  & \gray{60.84}      & \gray{44.84}  & \gray{48.75}  & \gray{41.12}  & \gray{68.51}  \\
\gray{SatMAE++~\cite{satmaepp}} & \gray{ViT-L}  & \gray{98.91}  & \gray{94.61}      & \gray{85.89}  & \gray{79.08}      & \gray{65.18}  & \gray{60.40}      & \gray{45.86}  & \gray{52.02}  & \gray{59.82}  & \gray{71.31}  \\ \bottomrule
\end{tabular}
    }
\end{table*}

\textbf{Classification Setup.}~We evaluate on three GFM-Bench classification datasets: EuroSAT \cite{eurosat}, BigEarthNet \cite{ben} and So2Sat \cite{so2sat}. For model evaluation, we attach a single-layer linear classification head to each encoder and employ two adaptation protocols: fine-tuning (FT), which updates all parameters, and linear probing (LP), which only trains the classification head while keeping the encoder frozen. Following prior work~\cite{croma, satmae}, we use 10\% of the training data for fine-tuning BigEarthNet and So2Sat. Models are fine-tuned for 20 epochs on EuroSAT and So2Sat, and 10 epochs on other benchmarks. LP experiments train for 100 epochs. For BigEarthNet, which is a multi-label classification task, we report mean Average Precision.

\textbf{Segmentation Setup.}~We evaluate our LESS ViT and baseline models on three of the GFM-Bench segmentation tasks: SegMunich~\cite{spectralgpt}, DFC2020~\cite{dfc}, and MARIDA~\cite{marida}. For downstream segmentation tasks, we adopt UPerNet~\cite{upernet} as the decoder and fine-tune the entire model end-to-end for 10 epochs.

\textbf{Results.}~Our experimental results for classification and segmentation tasks are summarized in \Cref{tab:quant}. LESS ViT demonstrates competitive performance across most benchmarks compared to existing approaches. Among ViT-Base-sized models, our approach achieves the second-highest average performance, indicating robust generalization across diverse downstream tasks. Notably, LESS ViT outperforms several ViT-Large-sized baselines on specific benchmarks, despite its more compact architecture. The linear probing (LP) results demonstrate that LESS ViT learns transferable representations for hyperspectral geospatial data, while the fine-tuning (FT) results highlight its strong task-specific adaptation capabilities. However, we observe performance gaps on So2Sat\cite{so2sat} and DFC2020~\cite{dfc} datasets. These gaps arise from distribution shifts between training and test sets, motivating future work to enhance training robustness through improved self-supervision objectives.

\begin{table}[t]
    \centering
    \caption{\textbf{Top: Cross-Satellite Generalization to Landsat.} We evaluate architectures for cross-satellite generalization on 20-channel Landsat segmentation. Models are fine-tuned on NLCD-L~\cite{ssl4eo-l}, with performance measured by mIoU. \textbf{Top and Bottom: Model Efficiency.} We analyze computational efficiency by comparing encoder parameter counts, floating point operations (FLOPs) during fine-tuning, and both fine-tuning and inference latency on NLCD-L (20 channels) and BigEarthNet~\cite{ben} (12 channels). Performance is reported as mIoU for NLCD-L and mAP for BigEarthNet. To enable direct comparison, we normalize both FLOPs and wall-clock times relative to LESS ViT's baseline measurements.}
    \label{tab:landsat}
    \footnotesize
    \renewcommand{\arraystretch}{1.2} 
    \resizebox{\textwidth}{!}{
    \begin{tabular}{llcccccc}
        \toprule
        \textbf{Dataset} & \textbf{Architecture}                          & \textbf{Backbone}      & \textbf{\#Param.}  & \textbf{Fine-Tuning Time}  & \textbf{\# FLOPs} & \textbf{Inference Time}& \textbf{Metric}         \\ \midrule
        \multirow{3}{*}{NLCD-L~\cite{ssl4eo-l}} & SatMAE~\cite{satmae}            & ViT-B         & 86.1M     & $\times 0.3$  & $\times 0.6 $ & $\times 0.3$      & 18.05             \\
        & Channel-ViT~\cite{channel_vit}  & Channel-ViT-B         & 85.4M     & $\times 2.6$  & $\times 3.1 $ & $\times 3.6$    & 10.35             \\
        & \bluecell{Ours}                   & \bluecell{LESS ViT-B}    & \bluecell{83.2M}     & \bluecell{$\times 1.0$}  & \bluecell{$\times 1.0 $} & \bluecell{$\times 1.0$}    & \bluecell{\textbf{24.31}}             \\ \midrule 
        \multirow{3}{*}{BigEarthNet~\cite{ben}} & SatMAE~\cite{satmae}     & ViT-B     & 86.1M     & $\times0.2$   & $\times0.6$  & $\times 0.1$    &\textbf{87.57}\\
& Channel-ViT~\cite{channel_vit} & Channel-ViT-B & 85.4M     & $\times3.3$  & $\times2.4$ & $\times2.0$ & 80.00 \\
& \bluecell{Ours}     & \bluecell{LESS ViT-B} & \bluecell{83.2M}    & \bluecell{$\times1.0$} & \bluecell{$\times1.0$}   &\bluecell{$\times1.0$} & \bluecell{86.08}\\ \bottomrule
\end{tabular}
}
\end{table}
\subsubsection{Cross-Satellite Generalization}
To demonstrate the flexibility of our LESS ViT architecture in handling satellites with varying channel counts without architectural modifications, we evaluate our model on the NCLD-L dataset from GFM-Bench. We construct this dataset by combining optical data from Landsat 7 and Landsat 8-9 from SSL4EO-L~\cite{ssl4eo-l}, resulting in a 20-channel geospatial dataset that exceeds Sentinel-2's channel count. We compare LESS ViT against two baseline architectures: SatMAE~\cite{satmae}, a ViT-based model, and Channel-ViT~\cite{channel_vit}, which explicitly models spatial-spectral attention.

\textbf{Model Setup.}~Standard ViT models have fixed channel capacity, requiring architectural modifications for additional spectral bands. For SatMAE~\cite{satmae}, we adapt its patch embedding layer by increasing the channel dimension and initializing new weights through wavelength-based interpolation. While both LESS ViT and Channel-ViT~\cite{channel_vit} can handle arbitrary channel counts through Tied Patch Embedding, Channel-ViT's explicit spatial-spectral attention mechanism incurs higher computational costs. Following the original implementation, Channel-ViT requires random channel masking (50\% of the channels) for successful fine-tuning on 20-channel data. In contrast, LESS ViT directly generalizes to such data without requiring architectural modifications or data manipulation. 

\textbf{Results.}~We fine-tune the three Sentinel-pretrained base models on NCLD-L for 10 epochs, with results shown in \Cref{tab:landsat}. Beyond channel count differences, Landsat features lower spatial resolution (30.0 meters/pixel) compared to Sentinel (10.0 meters/pixel), as shown \Cref{table:geo-bench}. From the results we can see that our model architecture generalizes to these variations better than the previous architectures.

\subsection{Qualitative Evaluation}\label{sec: qual_res}
\begin{figure*}[t]
    \centering
    \includegraphics[width=0.88\textwidth]{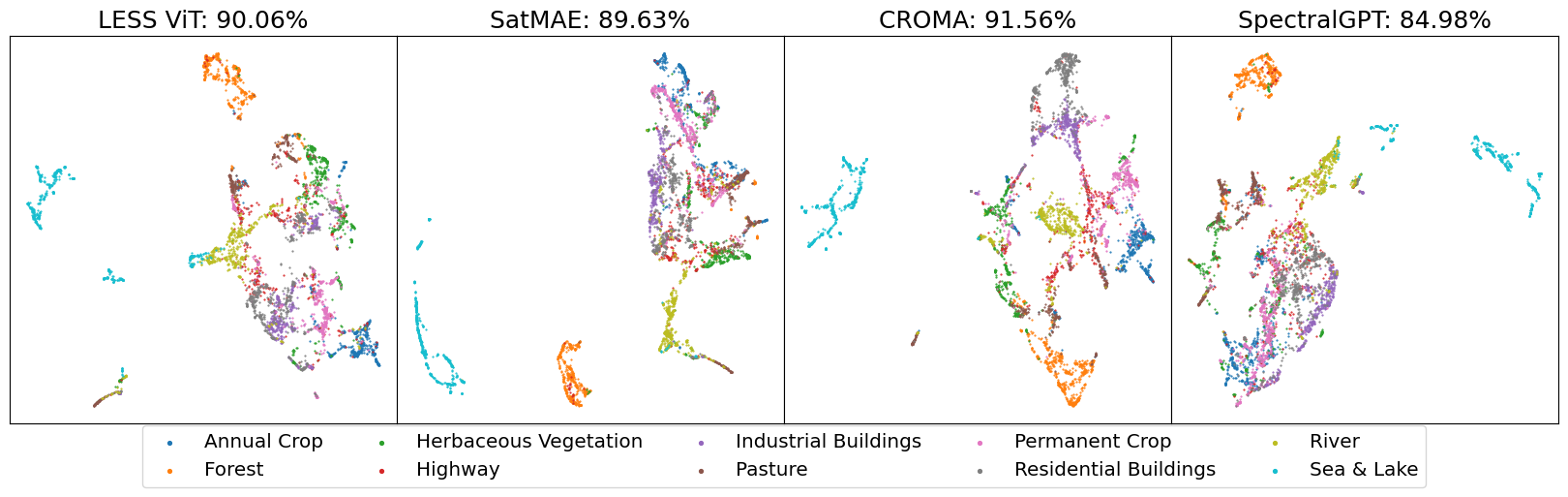}
    \caption{UMAP embedding and $k$NN accuracy ($k=20$) of our LESS ViT comparing with SatMAE~\cite{satmae}, CROMA~\cite{croma} and SpectralGPT~\cite{spectralgpt} models on EuroSAT~\cite{eurosat}.}
    \label{fig:umap}
\end{figure*}
\textbf{Clustering of Class Features.}~We visualize the UMAP \cite{umap} embeddings of the \texttt{[CLS]} tokens extracted by our pretrained model on the EuroSAT~\cite{eurosat} dataset and compare the results with those of previous works \cite{satmae, croma, spectralgpt}. Since \texttt{[CLS]} tokens serve as global representations of the entire image, samples from the same class should cluster together while different classes should remain distinct in the UMAP visualization. As shown in Figure \ref{fig:umap}, our method achieves clear class separation, indicating that our LESS ViT learns discriminative features that effectively capture class-specific characteristics. To further evaluate the quality of the extracted features, we perform $k$-nearest neighbor ($k$NN) classification and report the accuracy. In conclusion, the performance of our model, without any additional fine-tuning, demonstrates the effectiveness of our model architecture in capturing meaningful representations from geospatial data.

\begin{figure*}[t]
    \centering
    \includegraphics[width=\textwidth]{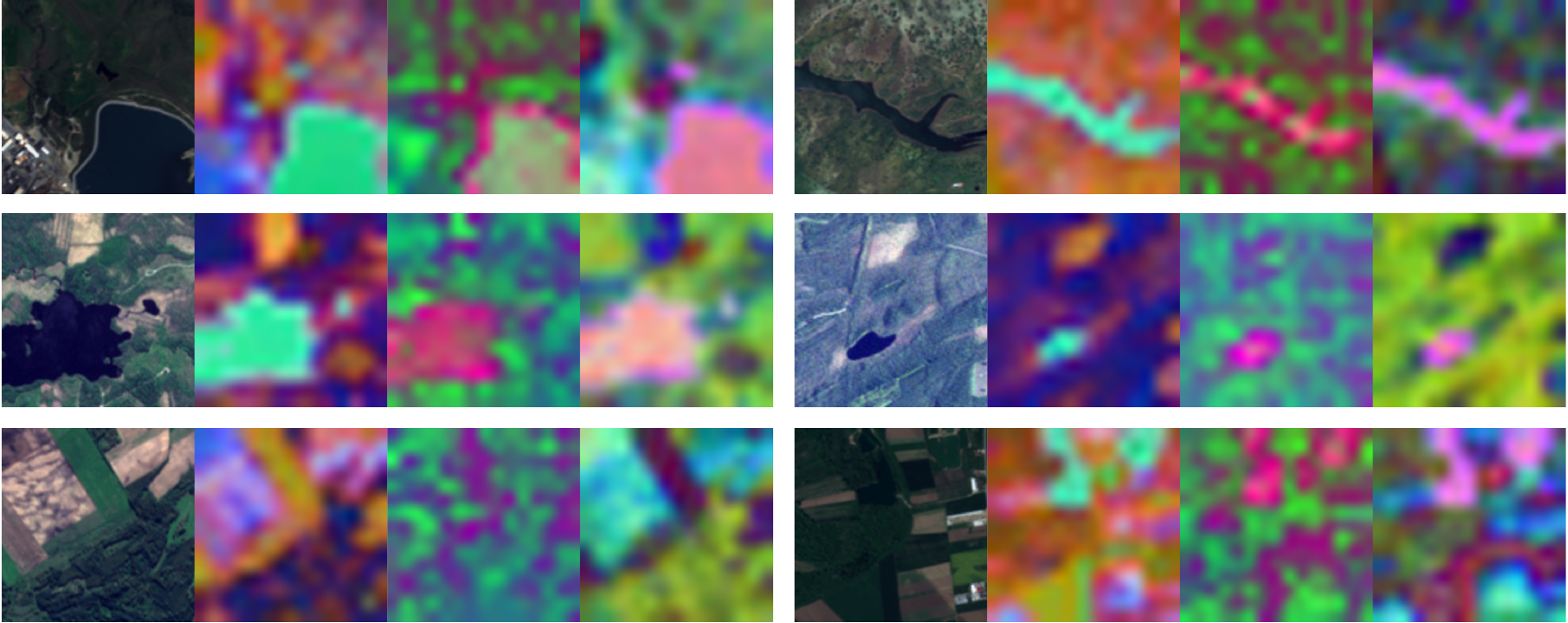}
    \caption{\textbf{Visualization of top PCA components on BigEarthNet~\cite{ben}}. Features are extracted using LESS ViT followed by PCA visualization. The four columns show the RGB image and visualizations for optical, radar, and multi-modal patch features respectively. Similar components are matched between images, indicating specific land cover and surface textures.} 
    \label{fig:pca}
\end{figure*}
\textbf{PCA of Patch Features.}~We perform principal component analysis (PCA) on the patch features extracted by LESS ViT. The top three principal components are visualized using three distinct colors, and the results are presented in \Cref{fig:pca}. We interpolate the resulting image back to the original input size for better interpretation. There are three key observations. First, the multi-modal patch features are mainly influenced by optical features, which is expected given the higher number of optical channels compared to radar channels. Second, for both the optical and multi-modal patch features, the principal components of similar land cover types have consistent color patterns. For instance, in optical feature visualizations, water bodies (Row 1) are highlighted in green, woodlands (Row 2) are represented by purple, and farmlands (Row 3) are primarily marked in yellow. Third, despite the presence of salt-and-pepper noise, the radar patch features demonstrate a correspondence between surfaces with similar textures. Smooth surfaces (water) tend to be marked in pink, while rough surfaces (woodlands and farmlands) are generally highlighted in green.

\section{Ablation Studies}
\subsection{Model Efficiency}
To evaluate LESS ViT's efficiency, we measure fine-tuning and inference wall-clock time, parameter counts, and floating point operations (FLOPs) compared to ViT~\cite{satmae} and Channel-ViT~\cite{channel_vit}. We compute fine-tuning and inference times on NLCD-L~\cite{ssl4eo-l} and BigEarthNet~\cite{ben}. All measurements except parameter counts are normalized relative to LESS ViT.

As shown in \Cref{tab:landsat}, both LESS ViT and Channel-ViT reduce parameter counts compared to standard ViT through their tied patch embedding layers, which share embedding weights across spectral channels. LESS ViT achieves the lowest parameter count through its low-rank attention module. ViT demonstrates the fastest fine-tuning and inference times by collapsing the spectral dimension during patch embedding, though it omits the spectral attention. In contrast, Channel-ViT's explicit spatial-spectral attention computation leads to the highest FLOPs and lowest computational efficiency. Despite computing spatial-spectral attention explicitly, Channel-ViT fails to benefit from this approach as it does not outperform standard ViT-based models, constrained by the inevitable random channel masking during training. Conversely, our LESS ViT approximates spatial-spectral attention more efficiently, eliminating the need for channel masking and enabling better training data utilization. This analysis aligns with our summary in \Cref{tab:summarize}, demonstrating that LESS ViT achieves better computational efficiency than Channel-ViT while preserving spectral attention capabilities.

\subsection{Model Design}
\begin{table*}[t]
    \centering
    \caption{\textbf{Ablation on LESS Attention Ratio ($\frac{d_1}{d_2}$) and Decoder Depth.} We evaluate the performance of LESS ViT across different model configurations on GFM-Bench. \textbf{Bold} values indicating the highest results.}
    \label{tab:abl}
    \footnotesize
    \renewcommand{\arraystretch}{1.2} 
    \setlength{\tabcolsep}{1pt} 
    \resizebox{\textwidth}{!}{
    \begin{tabular}{c@{\hspace{8pt}}
                    c@{\hspace{8pt}}
                    >{\centering\arraybackslash}p{1.6cm}
                    >{\centering\arraybackslash}p{1.6cm}
                    >{\centering\arraybackslash}p{1.6cm}
                    >{\centering\arraybackslash}p{1.6cm}
                    >{\centering\arraybackslash}p{1.6cm}
                    >{\centering\arraybackslash}p{1.6cm}
                    >{\centering\arraybackslash}p{1.6cm}
                    >{\centering\arraybackslash}p{1.6cm}
                    >{\centering\arraybackslash}p{1.6cm}
                    >{\centering\arraybackslash}p{1.6cm}
                    >{\centering\arraybackslash}p{1.6cm}
                    >{\centering\arraybackslash}p{1.6cm}}
        \toprule
                                &               & \multicolumn{2}{c}{EuroSAT}       & \multicolumn{2}{c}{BigEarthNet*}  & \multicolumn{2}{c}{So2Sat*}       & SegMunich     & DFC2020       & MARIDA        &               \\
                                &               & \multicolumn{2}{c}{Top 1 Acc.}    & \multicolumn{2}{c}{mAP}           & \multicolumn{2}{c}{Top 1 Acc.}    & mIoU          & mIoU          & mIoU          & \textit{Avg.} \\                  \cmidrule(lr){3-4} \cmidrule(lr){5-6}\cmidrule(lr){7-8}\cmidrule(lr){9-9}\cmidrule(lr){10-10}\cmidrule(lr){11-11}
Ratio                          & Decoder Depth      & FT            & LP                & FT            & LP                & FT            & LP                & FT            & FT            & FT            &               \\                 
\multirow{2}{*}{16}             & 4             & \textbf{97.98}         & 94.56             & 85.67         & 82.02             & 61.43         & \textbf{63.55}             & 41.03         & \textbf{48.90}         & 50.31         & 69.49         \\
                                & 8             & 97.94         & \textbf{94.72}             & \textbf{85.71}         & 81.81             & 61.54         & 62.58             & 41.07         & 47.67         & \textbf{54.74}         & \textbf{69.75}         \\ \midrule
\multirow{2}{*}{64}             & 4             & 97.76         & 94.44             & 85.59         & 82.21             & 62.48         & 61.72             & 40.92         & 44.90         & 53.01         & 69.23         \\
                                & 8             & 97.74         & 94.22             & 85.66         & \textbf{82.23}             & \textbf{62.50}         & 60.82             & \textbf{41.21}         & 48.52         & \textbf{54.74}         & 69.74         \\
        \bottomrule
    \end{tabular}
    }
\end{table*}
We examine the effects of attention ratio ($\frac{d_1}{d_2}$) in LESS ViT and decoder depth in Hyper-MAE through this ablation study. Our experiments evaluate two attention ratios ($\frac{d_1}{d_2} \in {16, 64}$) and two decoder depths (4 and 8 blocks), resulting in four distinct configurations. Each model variant is pretrained for 100 epochs and evaluated on GFM-Bench. The results are shown in \Cref{tab:abl}.

Our analysis reveals two key findings. First, increasing decoder depth consistently improves model performance, aligning with observations from MAE~\cite{mae}. Second, the spectral attention mechanism demonstrates significant impact: models with attention ratio $\frac{d_1}{d_2}=16$ ($d_1=32$, $d_2=2$) outperform those with $\frac{d_1}{d_2}=64$ ($d_1=64$, $d_2=1$), suggesting that increased attention to the spectral dimension enhances model performance. This empirically validates our theoretical motivation for LESS ViT's design, confirming that spectral attention mechanisms are essential for effective hyperspectral geospatial data processing.

\subsection{Mixture-of-Experts (MoE) Classification}
In addition to FT and LP, we propose an MoE classification head for LESS ViT that leverages its unique spatial-spectral token structure. LESS ViT generates $C$ spatial \texttt{[CLS]} tokens (one per channel) and a global \texttt{[CLS]} token. Since each hyperspectral channel is sensitive to specific landcover types, and spatial \texttt{[CLS]} tokens capture channel-specific spatial features, we implement multiple linear layers as experts with gating functions. Each expert selects the top-$k$ informative channels for prediction, with the final prediction determined by majority voting. In this study, we evaluate how increasing the number of experts affects linear probing performance on both optical and multi-modal BigEarthNet~\cite{ben} datasets. The left plot of \Cref{fig:moe} illustrates the performance progression from traditional linear probing (zero experts) through increasing expert counts. The results show consistent performance gains with increasing expert count, demonstrating the effectiveness of leveraging spatial \texttt{[CLS]} tokens through the MoE framework for classification tasks.

\subsection{Attention Rank}
\begin{figure}
    \centering
    \includegraphics[width=\linewidth]{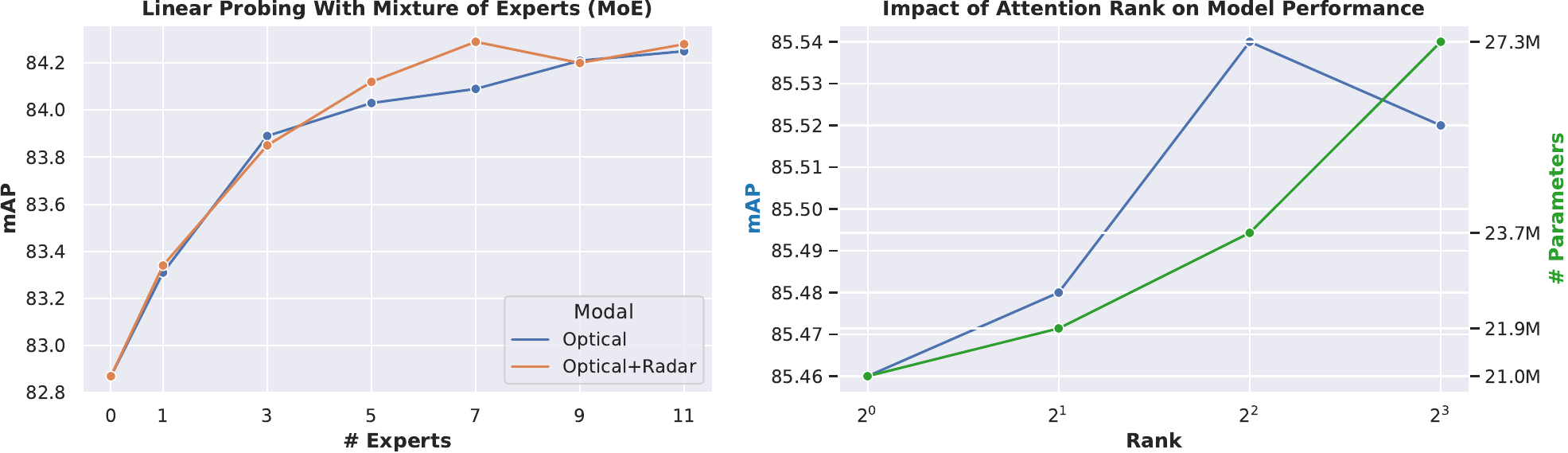}
    \caption{\textbf{Left: Linear Probing Results of Spectral Mixture of Experts (MoE) on BigEarthNet~\cite{ben}.} We evaluate the model performance through increasing numbers of experts, with each expert selects the top-3 informative channels to make the prediction. Experiments on both optical-only and optical-radar datasets demonstrate that our channel-wise MoE classification head enhances linear probing effectiveness. \textbf{Right: Finetuning Results of LESS ViT-S Models with Increasing Attention Ranks on BigEarthNet.} We evaluate performance and parameter count for LESS ViT-S models with attention ranks $r\in\{1, 2, 4, 8\}$. Results show that increasing rank leads to higher parameter counts and generally improves model performance.}
    \label{fig:moe}
\end{figure}

In this study, we examine the impact of the rank control hyperparameter $r$ by pretraining LESS ViT-Small models with $r\in\{1,, 2, 4, 8\}$. The LESS ViT-Small models have an embedding dimension of 384 and 6 attention heads, with an encoder depth of 12 and a decoder depth of 4. All other configurations remain consistent with the LESS ViT-Base experiments. We pretrained these models on the SSL4EO~\cite{ssl4eo} dataset for 100 epochs and fine-tuned them on the BigEarthNet~\cite{ben} dataset following our previous setup. Results are presented in the right plot of \Cref{fig:moe}.

The results demonstrate that as rank increases, the number of model parameters grows, generally leading to improved performance. The model achieves optimal performance at $r=4$, followed by a performance decline at $r=8$. This decline may be attributed to an insufficient number of epochs used in pretraining. Overall, model performance improves with increasing rank and gradually saturates when the rank becomes sufficiently large.

\begin{table}[t]
    \centering
    \caption{\textbf{Multi-Modal Performance Analysis.} We report the mAP scores for BigEarthNet~\cite{ben}, with the performance gains from radar modality integration shown in parentheses. Parameter counts for each multi-modal encoder are also provided for comparison. Following our evaluation protocol, we fine-tune on a 10\% subset of the BigEarthNet dataset. }
    \label{tab:mm}
    \footnotesize
    \renewcommand{\arraystretch}{1.2} 
    \setlength{\tabcolsep}{1pt} 
    \resizebox{0.5\textwidth}{!}{
    \begin{tabular}{c
                    c
                    c
                    >{\centering\arraybackslash}p{2.4cm}
                    >{\centering\arraybackslash}p{2.4cm}}
        \toprule
                                &               &                & \multicolumn{2}{c}{BigEarthNet*}   \\
                                &               &                & \multicolumn{2}{c}{mAP}            \\ \cmidrule(lr){4-5}\
Method                          & Backbone      & \# Param.      & FT            & LP                 \\
CROMA~\cite{croma}              & ViT-B         & 194.4M         & 87.87 (+0.30) & 85.13 (+0.23)      \\
\bluerow Ours                   & LESS ViT-B    & 83.2M          & 86.30 (+0.22) & 82.94 (+0.00)    \\ \bottomrule
\end{tabular}
}
\end{table}
\subsection{Multi-Modal Experiments}
BigEarthNet~\cite{ben} provides spatially aligned Sentinel-1 SAR and Sentinel-2 MSI data, enabling evaluation of LESS ViT's multi-modal capabilities. We compare our model with CROMA~\cite{croma}, the only multi-modal method among our baseline approaches. Following our single-modal evaluation pipeline, we present the multi-modal performance results and corresponding gains from radar channel integration in \Cref{tab:mm}. LESS ViT achieves comparable multi-modal performance gains (reported in the parenthesis) to the state-of-the-art method on fine-funing but on linear probing. This difference arises because CROMA uses separate encoders for different modalities plus a fusion encoder, doubling the model size compared to LESS ViT and yielding better zero-shot multi-modal features. In contrast, LESS ViT employs a single backbone for both single and multi-modal inputs, causing zero-shot features to be dominated by optical input. However, after fine-tuning, the performance gains become comparable, demonstrating LESS ViT's parameter efficiency in multi-modal information processing. Nevertheless, improving the multi-modal fusion capabilities for our LESS ViT architecture remains an important direction for our future work.

\section{Conclusion}
In this work, we present LESS ViT, a novel and flexible architecture for geospatial foundation model, which achieves superior performance with less computation and fewer parameters. The architecture incorporates three key components: the Hyperspectral Patch Embedding Block, the Low-rank Efficient Spatial-Spectral Attention Block, and the Perception Field Mask. These modules preserve the physical properties of geospatial data while ensuring efficiency and generalization across arbitrary image resolutions and channel counts. For pretraining, we develop Hyper-MAE, which extends the MAE paradigm by implementing decoupled spatial and spectral masking to create a challenging self-supervised learning objective. To enable systematic evaluation, we introduce GFM-Bench, a comprehensive benchmark suite that provides standardized validation sets for hyperparameter tuning across multiple datasets. This unified evaluation protocol facilitates fair model comparison and reproducible experimentation.

While LESS ViT demonstrates competitive performance with improved computational efficiency, extending the approach to extra dimensions (e.g., temporal dimension) remains a challenge. Although LESS attention can accommodate additional dimensions, this expansion results in reduced embedding dimensions $d_n$ per dimension. We propose exploring model scaling strategies through enhanced embedding dimension allocation in future research. Additionally, our current approach is limited to raster representations, while the remote sensing community possesses rich domain knowledge typically represented in vector formats, such as digital elevation models and slope models. Integrating these vector-based domain knowledge with raster (imagery) data remains an unresolved research challenge for future methodological advancements. Nevertheless, this work advances architectural design for hyperspectral data processing and deepens our understanding of multidimensional correlations. The proposed framework establishes a foundation for future developments in Earth Observation tasks and geospatial data analysis in the remote sensing community.

\section*{Acknowledgements}
This work was supported by a research grant from IBM-IL Discovery Accelerator Institute (IIDAI), an NSF NAIRR grant number NAIRR240419, and an NVIDIA's academic grants program. HZ was partially supported by a Google Research Scholar Award.

\printbibliography
\clearpage
\setcounter{page}{1}
\appendix
\section{Vanilla Spatial-Spectral Attention Blocks}
Channel-ViT~\cite{channel_vit} employs a straightforward but computationally inefficient approach to applying the attention mechanism to spatial-spectral tokens. It flattens the tokens into a matrix $\bar X\in\mathbb R^{NC \times D}$ and directly feeds $\bar X$ into standard attention blocks. However, the computational complexity of this approach grows quadratically with respect to the number of channels and tokens, which can be prohibitively expensive for large-scale applications. \Cref{algo:spatial-spectral} presents the specific algorithm used for the spatial-spectral attention blocks in Channel-ViT. The resulting computation complexity is $O(N^2C^2D)$. Meanwhile, our low-rank spatial-spectral attention block has a computation complexity of $O(N^2d_1+C^2d_2+NCD)$.

\setlength{\textfloatsep}{5pt}
\begin{algorithm}[h]
    \caption{Spatial-Spectral Attention Block~\cite{channel_vit}}\label{algo:spatial-spectral}
    \begin{algorithmic}[1]
        \Require Input tokens $X \in \mathbb{R}^{N \times C \times D}$
        \Ensure Output tokens $Y \in \mathbb{R}^{N \times C \times D}$
        \State $\bar{X} \gets \textsc{Reshape}(X, (NC, D)) $ \Comment{Flatten input tokens}
        \State $Q, K, V \gets \bar{X}W_Q, \bar{X}W_K, \bar{X}W_V \in \mathbb{R}^{NC \times D}$
        \State $Y \gets \textsc{Softmax}(QK^\top/\sqrt{d_k})V\in \mathbb{R}^{NC \times D}$ \Comment{Complexity $O(N^2C^2D)$}
        \State $Y \gets \textsc{Reshape}(Y, (N, C, D))$
        \State \Return $Y$
    \end{algorithmic}
\end{algorithm}

\section{pretraining Details} \label{appendix:pretrain}
\subsection{Data}
We pretrain our model using the SSL4EO-S12 dataset~\cite{ssl4eo}, which is a large-scale multi-modal dataset that consists of spatially and temporally aligned SAR and MSI images from Sentinel-1 GRD and Sentinel-2 L2A spectral satellites. Each SAR image includes VV and VH backscatter polarizations, while each MSI image provides 13 surface reflectance multispectral bands. The dataset covers 250K non-overlapping locations worldwide, with each location providing images from four seasons. The spatial resolution of the images is 2.64km $\times$ 2.64km. During pretraining, we randomly sample one optical-radar pair from the four seasons for each location at each epoch. For further details about the dataset, please refer to the SSL4EO paper~\cite{ssl4eo}.

\textbf{Channel Index.}~As we index the channels by the wavelength of the corresponding bands, we present the wavelength of each band in Sentinel-2 in \Cref{tab:channel_id}. All the presented experiments are developed based on Sentinel-2 data, and their channel indices can be referred to from the table.
\begin{table*}[!h]
    \centering
    \caption{\textbf{The wavelength (nm) of each band in Sentinel-2.} We select the corresponding wavelength as the channel index used in our hyperspectral patch embedding.}
    \label{tab:channel_id}
    \renewcommand{\arraystretch}{1} 
    \resizebox{\textwidth}{!}{
\begin{tabular}{ccccccccccccc}
\toprule
    \texttt{B0} &     \texttt{B1} (B) &     \texttt{B2} (G) &     \texttt{B3} (R) &     \texttt{B4} &     \texttt{B5} &     \texttt{B6} &     \texttt{B7} &      \texttt{B8} &     \texttt{B9} &     \texttt{B10} &     \texttt{B11} &     \texttt{B12} \\
\midrule
 442.7 &  492.4 &  559.8 &  664.6 &  704.1 &  740.5 &  782.8 &  832.8 &   864.7 &  945.1 &  1373.5 &  1613.7 &  2202.4 \\
\bottomrule
\end{tabular}
}
\end{table*}
\subsection{Implementation}
We train our main model for 300 epochs on 8$\times$NVIDIA A6000 GPUs, each with 48G VRAM, which takes 72 hours. The model is trained with an effective batch size of 1024, bfloat16 precision, a base learning rate of 1.5e-4, warmup for 5\% of the total epochs, and cooldown via a cosine annealing scheduler. We optimize the model using AdamW~\cite{adamw} optimizer with a weight decay of 5e-2. We normalize data following previous works~\cite{satmae, croma}. For data augmentation, we first randomly crop an area of $128 \times 128$ pixels from the original image with $264 \times 264$ pixels. Then, we apply the random horizontal flip with a probability of 50\%. Importantly, identical transformations are applied to both modalities to ensure spatial alignment. We do not resize the image to preserve the physical property of spatial resolution within the geospatial data.

\section{Evaluation Details}~\label{appendix:evaluation}
We evaluate our pretrained LESS ViT models and other baseline models on our proposed GFM-Bench. Following GEO-Bench, we adopt a standard approach by reserving 10\% of the training set as the validation set, selecting models and hyperparameters based on validation performance, and reporting the results on the test set.mOur pretrained and baseline models are evaluated through fine-tuning and linear probing experiments on classification tasks, as well as fine-tuning on segmentation tasks. As we are developing geospatial foundation models, off-the-shelf ability is crucial. Specifically, we aim to adapt the models to downstream tasks as efficiently as possible. Therefore, we limit fine-tuning to 10 epochs for larger datasets and 20 epochs for smaller datasets. For linear probing experiments, since the linear head updates quickly and requires fewer computational resources, we set all linear probing epochs to 100.

\textbf{Data Normalization.}~Following SatMAE~\cite{satmae}, we normalize the data by using the mean and standard deviation statistics of the entire dataset to clip out the pixel values that fall outside the 3\%-97\% range of the data distribution and normalize the data into range $(0,255)$ right before feeding data into the model. 

\subsection{Implementation}
\textbf{Fine-tuning.}~As different models and benchmark datasets may have varying optimal learning rates, we conduct a comprehensive search over a wide range of learning rates: \{3e-5, 5e-5, 8e-5, 1e-4, 3e-4, 5e-4, 8e-4, 1e-3\}. This search is applied to all models and benchmark datasets mentioned above, and we report the best results achieved across all learning rates. During fine-tuning, all models use an effective batch size of 256 and bfloat16 precision. The AdamW~\cite{adamw} optimizer is employed for all models with a weight decay of 1e-2, $\beta_{1}=0.9$, and $\beta_{2}=0.999$. Additionally, we utilize a cosine annealing scheduler with a warmup period last for 20\% of total fine-tuning epochs to adjust the learning rate during training.

We apply same data augmentations to the classification and segmentation of downstream tasks during training. Specifically, we start from normalizing all the images. Then we apply random flips both horizontally and vertically, as well as a random rotation. We still avoid any resize operations during data preprocessing following the data augmentation strategies during pretraining. For evaluation, we only normalize the input data. When reproducing baseline results, we apply appropriate resizing operations to accommodate different model input resolutions.

\textbf{Linear Probing.}~During the linear probing of image features, same data preprocessing pipeline is applied. We perform a comprehensive sweep of learning rates: \{5e-3, 8e-3, 1e-2, 3e-2, 5e-2, 8e-2, 1e-1, 3e-1\}. The training is conducted using an effective batch size of 1024 and bfloat16 precision. We employ the AdamW optimizer with the following hyperparameters: $\beta_{1}=0.9$ and $\beta_{2}=0.999$. We also set the weight decay to 1e-2. To manage the learning rate throughout the training process, we use a cosine annealing scheduler. The linear probing is carried out for a total of 100 epochs, which includes a warmup period of 20 epochs.
\end{document}